%% file: root.tex
\documentclass[letterpaper, 10 pt, conference]{ieeeconf}  % Comment this line out if you need a4paper

\IEEEoverridecommandlockouts                              % This command is only needed if 
\usepackage{amsmath} % assumes amsmath package installed
\usepackage{amssymb}  % assumes amsmath package installed
\usepackage{graphicx}
\usepackage{caption}
\usepackage{booktabs}
\usepackage{float}
\usepackage{cite}
\usepackage{color}
\usepackage[table]{xcolor}
\usepackage{multirow}
\usepackage{fontawesome5}
\usepackage{algorithm}
\usepackage{algorithmic}
\usepackage{hyperref}
\usepackage{fancyhdr}
\definecolor{darkgreen}{RGB}{34,139,34}

\title{\LARGE \bf
VertiAdaptor: Online Kinodynamics Adaptation for \\ Vertically Challenging Terrain
}

\author{Tong Xu, Chenhui Pan, Aniket Datar, and Xuesu Xiao% <-this % stops a space
\thanks{All authors are with the Department of Computer Science, George Mason University {\tt\small \{txu25, cpan7, adatar, xiao\}@gmu.edu}}% <-this % stops a space
}

\begin{document}

\maketitle
\thispagestyle{empty}
\pagestyle{empty}
% \thispagestyle{withfooter}
% \pagestyle{withfooter}

\input{Contents/abstract}
\input{Contents/intro}

\input{Contents/related_work}
\input{Contents/method}

\input{Contents/implement}
\input{Contents/results}

\input{Contents/conclusion}


%%%%%%%%%%%%%%%%%%%%%%%%%%%%%%%%%%%%%%%%%%%%%%%%%%%%%%%%%%%%%%%%%%%%%%%%%%%%%%%%

%%%%%%%%%%%%%%%%%%%%%%%%%%%%%%%%%%%%%%%%%%%%%%%%%%%%%%%%%%%%%%%%%%%%%%%%%%%%%%%%
\bibliographystyle{IEEEtran}
\bibliography{IEEEabrv,references}

% \begin{thebibliography}{99}

% \bibitem{c1} M. D. Teji, T. Zou, and D. S. Zeleke, “A survey of off-road mobile robots: Slippage estimation, robot control, and sensing technology,” \textit{Journal of Intelligent \& Robotic Systems}, vol. 109, p. 38, Oct 2023.
% \bibitem{c2} J. Schulman, F. Wolski, P. Dhariwal, A. Radford, and O. Klimov, “Proximal policy optimization algorithms,” \textit{arXiv preprint arXiv:1707.06347}, 2017.
% \bibitem{c3} A. Tasora, R. Serban, H. Mazhar \textit{et al}, “Chrono: An open source multi-physics dynamics engine,” in \textit{High Performance Computing in Science and Engineering: Second International Conference}, pp. 19-49, 2015.
% \bibitem{c4} A. Datar, C. Pan, M. Nazeri and X. Xiao, “Toward wheeled mobility on vertically challenging terrain: Platforms, datasets, and algorithms,” \textit{arXiv preprint arXiv:2303.00998}, 2023.
% \bibitem{c5} S. Narvekar, B. Peng, M. Leonetti, J. Sinapov, M. E. Taylor and P. Stone, “Curriculum learning for reinforcement learning domains: A framework and survey,”. \textit{Journal of Machine Learning Research}, pp. 1-50, 2020.
% \bibitem{c6} S. Kolouri, C. E. Martin, and G. K. Rohde, “Sliced-wasserstein autoencoder: an embarrassingly simple generative model,” \textit{arXiv.1804.01947}, 2018.

% \end{thebibliography}

\end{document}

%% file: Contents/abstract.tex
%%%%%%%%%%%%%%%%%%%%%%%%%%%%%%%%%%%%%%%%%%%%%%%%%%%%%%%%%%%%%%%%%%%%%%%%%%%%%%%%
\begin{abstract}

Autonomous driving in off-road environments presents significant challenges due to the dynamic and unpredictable nature of unstructured terrain. Traditional kinodynamic models often struggle to generalize across diverse geometric and semantic terrain types, underscoring the need for real-time adaptation to ensure safe and reliable navigation. We propose \textit{VertiAdaptor} (VA), a novel online adaptation framework that efficiently integrates elevation with semantic embeddings to enable terrain-aware kinodynamic modeling and planning via function encoders. VA learns a kinodynamic space spanned by a set of neural ordinary differential equation basis functions, capturing complex vehicle-terrain interactions across varied environments. After offline training, the proposed approach can rapidly adapt to new, unseen environments by identifying kinodynamics in the learned space through a computationally efficient least-squares calculation. We evaluate VA within the Verti-Bench simulator, built on the Chrono multi-physics engine, and validate its performance both in simulation and on a physical Verti-4-Wheeler platform. Our results demonstrate that VA improves prediction accuracy by up to 23.9\% and achieves a 5X faster adaptation time, advancing the robustness and reliability of autonomous robots in complex and evolving off-road environments.

\end{abstract}

%% file: Contents/intro.tex
%%%%%%%%%%%%%%%%%%%%%%%%%%%%%%%%%%%%%%%%%%%%%%%%%%%%%%%%%%%%%%%%%%%%%%%%%%%%%%%%
\section{Introduction}

Off-road autonomous navigation represents one of the most challenging frontiers in robotics, where vehicles must traverse complex, unstructured environments characterized by varying terrain properties, obstacles, and dynamic conditions. State-of-the-art off-road perception and planning systems often degrade under distribution shift inevitable in real-world enviornments~\cite{min2024autonomous, wang2024survey}. As a result, reliable off-road navigation requires both enhanced terrain understanding and online kinodynamics adaptation to accommodate changing vehicle-terrain interactions\cite{xiao2022motion}.

To improve terrain awareness, many systems augment geometric maps with semantics information. Elevation mapping has emerged as an efficient geometric representation for ground robots, with mature real-time implementations available\cite{miki2022elevation, fankhauser2018probabilistic, mattamala2025wild}. In parallel, new off-road datasets and platforms enable robust terrain semantic segmentation that extends beyond urban domains\cite{triest2022tartandrive}. Recent advances in learning-based traversability estimation\cite{cai2025pietra, elnoor2024pronav, jung2024v} and vision models that couple geometry with semantics (e.g., bird's-eye view traversability\cite{castro2022does} and multi-head terrain prediction\cite{meng2023terrainnet}) further strengthen the integration of elevation and semantics in off-road settings. Nevertheless, unifying elevation and semantics with real-time adaptation within a single framework remains an open challenge, due to the large space and variability of the elevation and semantic input and limited onboard computation. 

\begin{figure}[t!]
    \centering
    \includegraphics[width=\columnwidth]{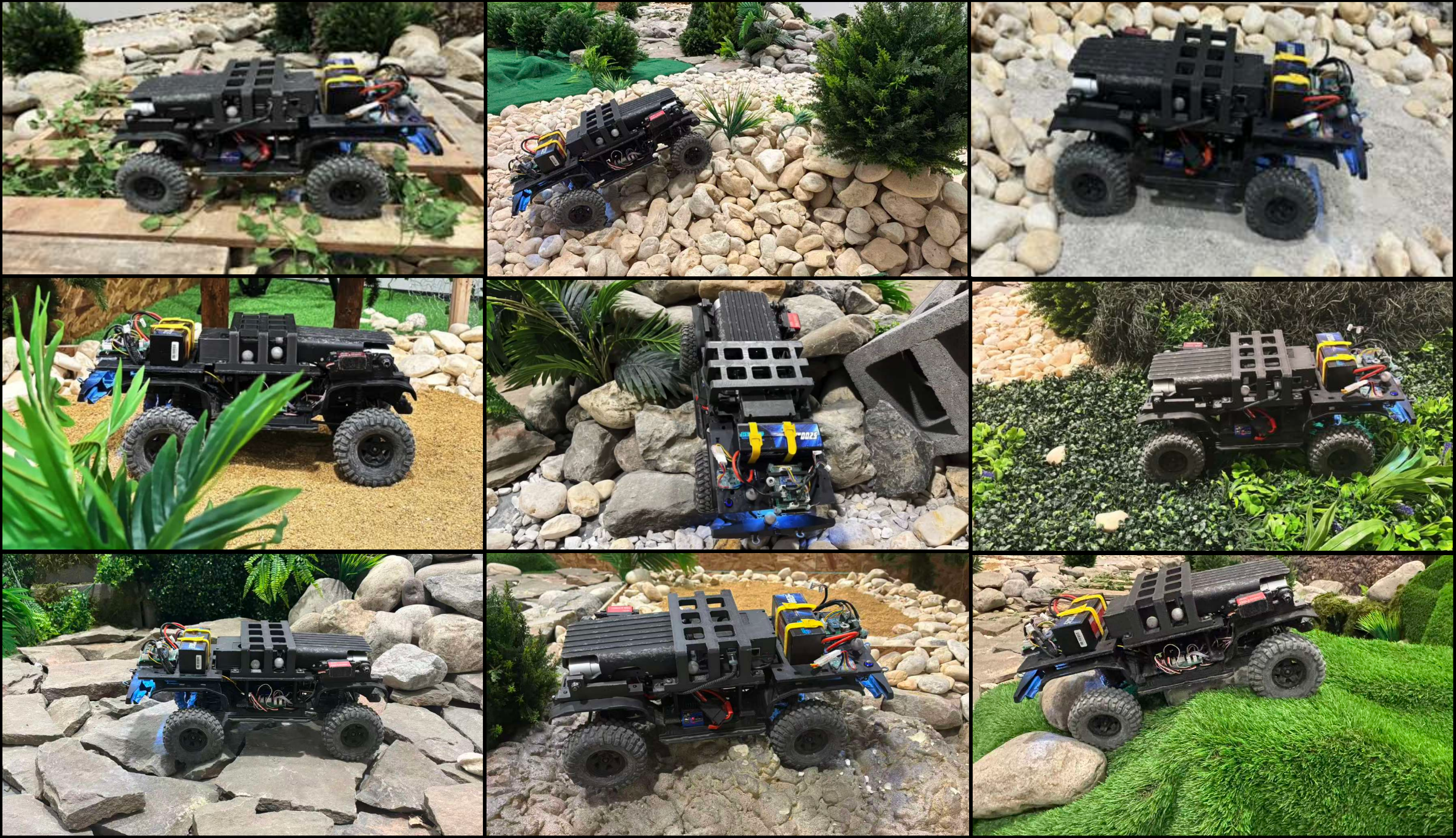}
    \caption{Facing a variety of constantly changing elevation and semantics on vertically challenging, off-road terrain, autonomous mobile robots need to quickly adapt their terrain understanding through online kinodynamics adaptation to achieve safe and efficient navigation.} 
    \label{fig::motivation}
    \vspace{-10pt}
\end{figure}

To address this limitation, meta-learning offers a promising path toward rapid online adaptation. Meta-trained kinodynamic models can be quickly updated with a small amount of recent data to track changes such as new terrain or partial system failures\cite{clavera2018model}. Recent advances in continuous-time neural modeling, particularly neural Ordinary Differential Equation (ODE), provide flexible function classes for system dynamics while retaining ODE structure for integration and control\cite{chen2018neural}. When combined with neural ODE, function encoders have emerged as a powerful technique to learn compact representations of complex function spaces\cite{ingebrand2024zero}. These methods enable efficient adaptation to new scenarios by parameterizing models within a learned basis function space, allowing for rapid online adaptation without extensive recomputation.

To push the boundaries of off-road wheeled mobility on diverse and evolving vertically challenging terrain (Fig.~\ref{fig::motivation}), we present \textbf{V}erti\textbf{A}daptor (VA), a novel online adaptation framework that efficiently integrates elevation with semantic embeddings for terrain-aware kinodynamic modeling. Our contributions can be summarized as follows:

\begin{figure*}[ht]
    \centering
    \includegraphics[width=2\columnwidth]{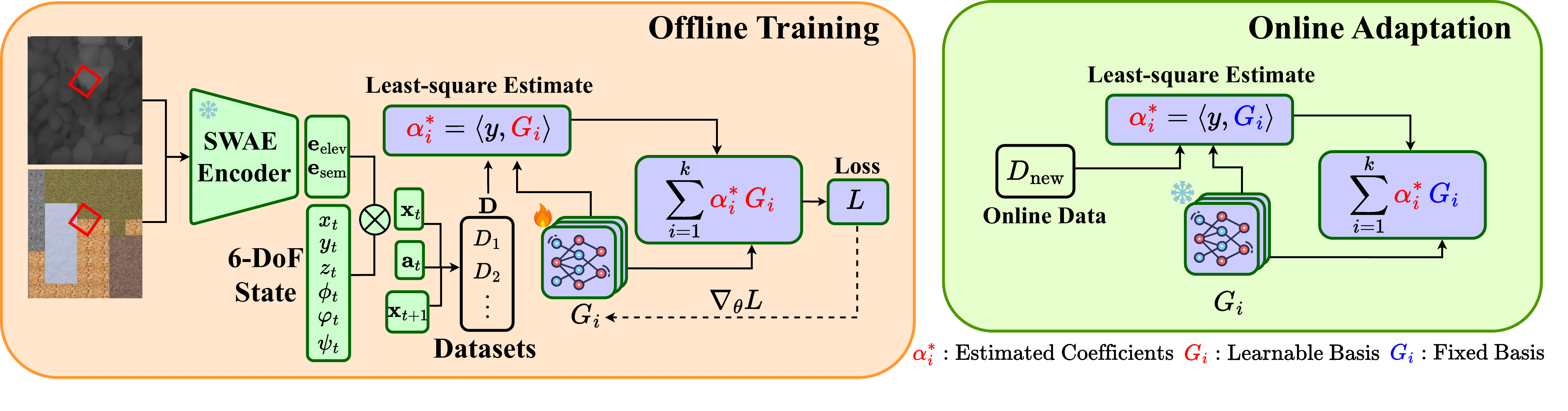}
    \caption{VertiAdaptor Overview: The offline training phase combines elevation and semantic embeddings to train neural ODE basis functions, i.e., state change rates $\{g_1, g_2, \ldots, g_k\}$ integrated into state changes $\{G_1, G_2, \ldots, G_k\}$. The online adaptation phase uses a small set of new data to identify the coefficients, enabling kinodynamics to be represented as a linear combination of basis functions.}
    \label{VA}
    \vspace{-15pt}
\end{figure*}

\begin{itemize}
    \item A unified terrain representation using the Sliced Wasserstein Autoencoder (SWAE)\cite{kolouri2018sliced} to efficiently obtain elevation and semantic embeddings beneath the vehicle; 
    % based on datasets collected via sinusoidal random exploration across 100 elevation geometries and ten terrain classes (seven rigid, three deformable) in the Verti-Bench\cite{xu2025verti};
    \item VA, the first  $\mathbb{SE}(3)$ kinodynamics online adaptation framework based on complex off-road elevation and semantics that enables rapid kinodynamic model updates through efficient least-squares optimization (Fig.~\ref{VA});
    \item An extensive autoregressive evaluation of different adaptation techniques, including Multi-Layer Perceptron (MLP) fine-tuning, Model-Agnostic Meta-Learning (MAML)\cite{nagabandi2018learning}, neural ODE~\cite{chen2018neural}, and VA, for off-road kinodynamic modeling; and
    \item Empirical validation of our method in simulated environments and on a physical Verti-4-Wheeler platform~\cite{datar2024toward}, showcasing better performance over baselines. 
\end{itemize}

%% file: Contents/related_work.tex
\section{Related Work}
In this section, we review related work in off-road mobility using kinodynamic models, as well as online adaptation techniques to mitigate distribution shift. 

\subsection{Off-road Mobility}
Kinodynamic modeling has been fundamental in robot motion planning, as it explicitly accounts for both kinematics and dynamics~\cite{lavalle2001randomized}. However, classical models suffer from significant limitations when applied to diverse, real-world, off-road enviornments. The assumption of known, constant terrain parameters rarely holds in practice, and the models often fail to capture nonlinear effects such as dynamic loading, tire deformation, and complex multi-terrain interactions. Considering the limitations of classical approaches, learning-based methods for off-road mobility have emerged as a promising alternative. Prior work has explored a variety of strategies, such as learning end-to-end policies~\cite{pan2020imitation}, data-driven kinodynamic models~\cite{williams2017information, otsu2016autonomous}, parameter adaptation~\cite{o2022neural, wang2021apple, xiao2022appl}, and cost functions~\cite{cai2024evora, frey2023fast}. These approaches have improved generalization across terrain compared to purely classical methods. However, most approaches assume that the learned dynamics remain static after training, which limits their effectiveness in environments where vehicle–terrain interactions can change rapidly.

\subsection{Meta Learning in Robotics}
Meta-learning, or ``learning to learn'', has significant advancement in robotics as a means of enabling agents to adapt quickly to new tasks or environments with limited data. Early approaches, such as MAML~\cite{finn2019online}, demonstrate that policies and models can be initialized to adapt rapidly with just a few gradient updates. Such property is particularly useful for robotics where real-world data collection is expensive.

Beyond episodic meta-learning, online adaptation methods have been proposed to handle dynamic environments. Methods such as online meta-learning\cite{finn2019online} and gradient-based adaptation with memory~\cite{rusu2018meta, liu2021lifelong}  allow robots to continually update their models while retaining knowledge of past tasks~\cite{liu2021lifelong}. In off-road settings, where vehicle–terrain dynamics shift rapidly, efficient online adaptation remains critical for maintaining performance and safety.
However, gradient-based adaptation methods often suffer from computational overhead and convergence issues during onboard deployment, especially when facing complex kinodynamic changes due to varying off-road elevation and semantics. 

The concept of function encoders\cite{ingebrand2024zero} represents a promising direction for kinodynamic model adaptation. By learning a basis of function encoders that span the space of possible dynamics, these methods enable rapid adaptation through efficient linear combinations. Nevertheless, previous function encoder approaches are primarily designed for 2D space with motion constrained to $\mathbb{SE}(2)$, which no longer holds when confronting off-road environments, especially vertically challenging terrain. Extending to $\mathbb{SE}(3)$, VA also efficiently incorporates underlying elevation and semantic information, which are crucial for terrain-aware online adaptation.

%% file: Contents/method.tex
\section{Method}
We first formulate the problem of forward kinodynamic modeling for wheeled mobility on vertically challenging terrain. We then describe how to efficiently obtain elevation and semantic embeddings beneath the vehicle. Finally, we introduce our VA method which integrates these embeddings to enable rapid kinodynamic model updates via least-squares optimization.

\subsection{Problem Formulation}
\label{sec::pf}
Most ground navigation problems in robotics are defined in the planar state space $X \subset \mathbb{SE}(2)$, where only 2D position and heading are considered. However, off-road mobility on vertically challenging terrain requires six Degrees of Freedom (DoFs) and needs to extend the state space to $X \subset \mathbb{SE}(3)$. We define the robot state at time $t$ as
\[
\mathbf{x}_t = \big[x_t, y_t, z_t, \phi_t, \varphi_t, \psi_t\big] \in \mathbb{SE}(3),
\]
where $x_t, y_t, z_t$ represent the vehicle position in 3D, and $\phi_t, \varphi_t, \psi_t$ denote roll, pitch, and yaw. Control inputs $\mathbf{u}_t \in \mathcal{U}$ correspond to vehicle steering and speed.  

The vehicle’s kinodynamic behavior can be modeled in continuous time as an ODE: 
\[
\dot{\mathbf{x}}{_t} = f_{\theta}(\mathbf{x}_t, \mathbf{u}_t, \mathbf{e}_t),
\]
where $\mathbf{e}_t \in \mathcal{E}$ denotes the environment embedding that combines terrain elevation and semantic information beneath the vehicle. For compatibility with the Model Predictive Path Integral (MPPI)\cite{williams2017model} sampling-based control framework, we adopt a discretized form:
\[
\mathbf{x}_{t+1} - \mathbf{x}_t = \int_t^{t+\Delta t} f_\theta(\mathbf{x}(\tau), \mathbf{u}(\tau), \mathbf{e}(\tau)) \, d\tau,
\]
with fixed time step $\Delta t$. The function $f_\theta$ is parameterized by a set of neural ODEs~\cite{chen2018neural}. The goal is to approximate forward kinodynamic model $f_\theta$ that captures nonlinear vehicle--terrain interactions while supporting efficient online adaptation in unseen terrain.  

\subsection{Elevation \& Semantic Embeddings}
\label{sec::elev_sem}
Unlike on-road driving, the underlying elevation and semantics play a vital role in determining off-road kinodynamics on vertically challenging terrain. For example, a large boulder may cause rollover, while deformable sand may get the vehicle stuck. Therefore, our environment embedding $\mathbf{e}_t$ unifies geometric and semantic information
\[
\mathbf{e}_t = [\mathbf{e}_{\text{elev}}, \mathbf{e}_{\text{sem}}] \in \mathbb{R}^{d_e + d_s},
\]
where $\mathbf{e}_{\text{elev}} \in \mathbb{R}^{d_e}$ represents local elevation features beneath the vehicle, and $\mathbf{e}_{\text{sem}} \in \mathbb{R}^{d_s}$ encodes semantic terrain properties such as friction and deformability. In our implementation, elevation is represented as 2.5D elevation maps, while semantic information is provided through bird’s-eye view RGB images, both aligned with the vehicle’s heading. To facilitate efficient modeling and adaptation, we use SWAE~\cite{kolouri2018sliced} to project raw elevation and semantic maps into a compact latent space. 

\subsection{Function Encoder for Neural ODEs}
To create accurate kinodynamic models across diverse terrain, we represent the forward kinodynamics as a linear combination of $k$ neural ODE basis functions:
\begin{align*}
\mathbf{x}_{t+1} - \mathbf{x}_t 
&= \sum_{i=1}^{k} \alpha_i \int_{t}^{t+\Delta t} 
    g_i\big(\mathbf{x}(\tau), \mathbf{u}(\tau), \mathbf{e}(\tau)\big)\, d\tau \nonumber \\
&= \sum_{i=1}^{k} \alpha_i \, 
    G_i\big(\mathbf{x}_t, \mathbf{u}_t, \mathbf{e}_t;\theta_i\big),
\end{align*}
where $G_i(\cdot;\theta_i)$ are a set of learnable basis functions parameterized by neural ODEs and $\theta_i$, and $\boldsymbol{\alpha} = [\alpha_1, \alpha_2, \ldots, \alpha_k]^T \in \mathbb{R}^k$ are the coefficients that define the contribution of each basis function. $g_i$ is the basis function state change rate to be integrated into $G_i$. Each $G_i$ outputs the predicted state change. In practice, we approximate the integral $G_i$ using the fourth-order Runge-Kutta (RK4)\cite{tan2012general} numerical integrator.

When the vehicle encounters an unseen terrain, we collect a small buffer of recent trajectories:

\begin{equation*}
\mathcal{D}_{\text{new}} = \{(\mathbf{x}_l, \mathbf{u}_l, \mathbf{e}_l, \mathbf{x}_{l+1})\}_{l=1}^M,
\end{equation*}
where $M$ is the limited number of online samples. Given the pre-trained basis functions $\{G_i\}_{i=1}^k$, the online adaptation task becomes a least squares problem to find optimal coefficients $\boldsymbol{\alpha}^*$. The optimal coefficients $\boldsymbol{\alpha}^*$ can be computed efficiently in closed form as the solution:

\begin{equation}
\boldsymbol{\alpha}^* = \begin{bmatrix}
\langle G_1, G_1 \rangle & \cdots & \langle G_1, G_k \rangle \\
\vdots & \ddots & \vdots \\
\langle G_k, G_1 \rangle & \cdots & \langle G_k, G_k \rangle
\end{bmatrix}^{-1} \begin{bmatrix}
\langle y, G_1 \rangle \\
\vdots \\
\langle y, G_k \rangle
\end{bmatrix},
\label{eq:coef}
\end{equation}
where $G_i$ denotes the $i$-th basis function evaluated on the new data, $y$ is the target state change, and $\langle \cdot, \cdot \rangle$ represents the vector inner products. This least-squares solution enables rapid online adaptation with a maximum time complexity of $O(k^3)$, far more efficient than gradient-based retraining.

\subsection{VertiAdaptor (VA)}
\label{sec::va}
The VA framework consists of two components: 
(i) \textit{offline training}, where neural ODE basis functions are learned across diverse environments, and (ii) \textit{online adaptation}, where coefficients are updated repeatedly to adapt to new unseen terrain.

\begin{algorithm}[h]
\caption{Multi-Step Training for VA}
\label{alg::va}
\begin{algorithmic}[1]
\STATE \textbf{Input:} Training dataset $\mathcal{D}$, number of basis functions $k$, learning rate $\alpha$, prediction horizon $T_{\text{pred}}$, rollout step size $\Delta t$
\STATE \textbf{Output:} Neural ODE basis functions $G_1, G_2, \ldots, G_k$
\STATE Initialize each basis function state change rate $g_i$ as a neural network with parameters $\theta_i$, for $i = 1, \ldots, k$
\WHILE{not converged}
    \STATE $L = 0$
    \STATE Sample $F$ environments and $N$ trajectories per environment from $\mathcal{D}$
    \FOR{each environment $f \in \{1, \ldots, F\}$}
        \STATE $\mathcal{T}_{\text{ex}}^{(f)} \leftarrow$ first $N_{\text{ex}}$ trajectories
        \STATE $\mathcal{T}_{\text{query}}^{(f)} \leftarrow$ remaining $N_{\text{q}}$ trajectories
        \STATE Compute coefficients $\boldsymbol{\alpha}^{*(f)} = [\alpha_1^{*(f)}, \ldots, \alpha_k^{*(f)}]$ using $\mathcal{T}_{\text{ex}}^{(f)}$ via Eqn.~\eqref{eq:coef}
        \FOR{each query trajectory $n \in \mathcal{T}_{\text{query}}^{(f)}$}
            \FOR{each sampled rollout $s \in \{1, \ldots, S\}$}
                \STATE Initialize $\mathbf{x}_0^{\text{pred}} \leftarrow$ initial state of rollout $s$
                \STATE Let $[\mathbf{u}_0, \mathbf{u}_1, \ldots, \mathbf{u}_{T_{\text{pred}}-1}]$ be the control sequence for rollout $s$
                \FOR{$t = 1, \ldots, T_{\text{pred}}$}
                    \FOR{$i = 1, \ldots, k$}
                        \STATE $G_i(\cdot) \leftarrow \text{RK4}(g_i, \mathbf{x}_{t-1}^{\text{pred}}, \mathbf{u}_{t-1}, \Delta t)$
                    \ENDFOR
                    \STATE $\mathbf{x}_t^{\text{pred}} \leftarrow \mathbf{x}_{t-1}^{\text{pred}} + \sum_{i=1}^k \alpha_i^{*(f)} G_i(\cdot)$
                    \STATE $L \leftarrow L + \|\mathbf{x}_t^{\text{pred}} - \mathbf{x}_t^{\text{gt}}\|^2$
                \ENDFOR
            \ENDFOR
        \ENDFOR
    \ENDFOR
    \STATE $\theta = \{\theta_1, \ldots, \theta_k\} \leftarrow \theta - \alpha \nabla_\theta L$
\ENDWHILE
\end{algorithmic}
\end{algorithm}

\subsubsection{Offline Training}
We employ a multi-step rollout approach to learn basis functions. First, with fixed basis functions, coefficients are computed by an example set. Second, the computed coefficients are used on a query set to output state changes, which are compared against ground truth to compute loss and update basis functions. 

Algorithm \ref{alg::va} details this procedure. In each training iteration, we sample $F$ environments and $N$ complete trajectories per environment from the training dataset $\mathcal{D}$ (line 6). Each environment's data are split into (the first $N_{\text{ex}}$) example trajectories for coefficient computation and (the remaining $N_{\text{q}}$) query trajectories to learn basis functions (lines 8-9). 
For each environment, coefficients $\boldsymbol{\alpha}^{*(f)}$ are computed using ground truth example trajectories via Eqn.~\eqref{eq:coef} (line 10). $S$ multi-step rollouts  of length $T_{\text{pred}}$ then occur for each query trajectory $n$ in the nested loops (lines 11-23), where each neural ODE basis function $G_i$ output is computed via RK4 integration (line 17) and linearly combined using environment-specific coefficients (line 19). The training loss accumulates prediction errors across all rollout steps $t$, rollouts $s$, and query trajectories $n$ (line 20) and becomes
\begin{equation*}
\mathcal{L}
= \sum_{n=1}^{N_{\text{q}}} \sum_{s=1}^{S} \sum_{t=1}^{T_{\text{pred}}}
\big\| \mathbf{x}^{\text{pred}}_{t, s, n} - \mathbf{x}^{\text{gt}}_{t, s, n} \big\|_2^{2},
\end{equation*}
where $\mathbf{x}^{\text{pred}}_{t, s, n}$ and $\mathbf{x}^{\text{gt}}_{t, s, n}$ denote the predicted and ground truth state at time $t$ on rollout $s$ in trajectory $n$ respectively. This multi-step loss encourages accurate long-term predictions. After convergence, the basis functions are frozen and ready for online adaptation.
 
\subsubsection{Online Adaptation}

During deployment, VA continuously adapts the coefficients $\boldsymbol{\alpha}$ every 5 seconds using recent trajectories. This continuous adaptation mechanism enables VA to maintain accurate kinodynamics as terrain conditions evolve during navigation. The adapted coefficients are immediately used for forward prediction in the MPPI system, enabling real-time response to evolving terrain while preserving the generalization benefits of the pre-trained basis functions.

%% file: Contents/implement.tex
\section{Implementations}

In this section, we describe the implementation details of dataset collection, preprocessing, the kinodynamic model architecture within the VA framework, and other baselines designed for adaptation to vertically challenging terrain.

\begin{table}[h]
\centering
\caption{Hyperparameters for VA}
\label{tab:params}
\small
\begin{tabular}{ll}
\toprule
\textbf{Parameter} & \textbf{Value} \\
\midrule
\multicolumn{2}{l}{\textbf{\textit{Model Architecture}}} \\
Number of basis functions ($k$) & 24 \\
Hidden dimension & $\lfloor 64\sqrt{k} \rfloor = 313$ \\
Activation function & ReLU \\
Output dimension & 6 \\
\midrule
\multicolumn{2}{l}{\textbf{\textit{Training Configuration}}} \\
Learning rate & $10^{-3}$ \\
Optimizer & Adam \\
Learning rate scheduler & Cosine annealing \\
Total training steps & 1000 \\
\midrule
\multicolumn{2}{l}{\textbf{\textit{Data Sampling}}} \\
Environments per batch ($F$) & 5 \\
Trajectories per environment ($N$) & 10 \\
Example trajectories ($N_{\text{ex}}$) & 4 \\
Query trajectories ($N_{\text{q}}$) & 6 \\
Sequences per trajectory ($S$) & 2 \\
Prediction horizon ($T_{\text{pred}}$) & 8 \\
\midrule
\multicolumn{2}{l}{\textbf{\textit{Integration}}} \\
RK4 time step ($\Delta t$) & 0.1 \\
Example batch size & 256 \\
Regularization ($\lambda$) & $10^{-3}$ \\
\bottomrule
\end{tabular}
\vspace{-10pt}
\end{table}

\subsection{Datasets and Preprocessing}
Based on the Verti-Bench\cite{xu2025verti} simulator, we construct 100 off-road environments, each spanning 129m$\times$129m with 0.1m pixel resolution. The environments are evenly distributed across low, medium, and high elevation levels. Each environment contains seven rigid (grass, wood, gravel, dirt, clay, rock, and concrete) and three deformable terrain semantic classes (snow, mud, and sand). Rigid terrain is modeled with a normal distribution of friction coefficients and a fixed restitution coefficient of 0.01, while deformable terrain is assigned physics parameters (cohesive effect, soil stiffness, and hardening effect) at three levels of deformability: soft, medium, and hard.  

Elevation and RGB semantic patches are extracted from a 128$\times$128 pixel region beneath and aligned with the vehicle at 10 Hz. Elevation values are centered at the vehicle's current altitude. For feature extraction, we employ the SWAE generative model to embed elevation and semantic patches into a 64-dimensional latent space, preserving both geometric and semantic information for downstream kinodynamic modeling.

\begin{figure*}[ht]
    \centering
    \includegraphics[width=2\columnwidth]{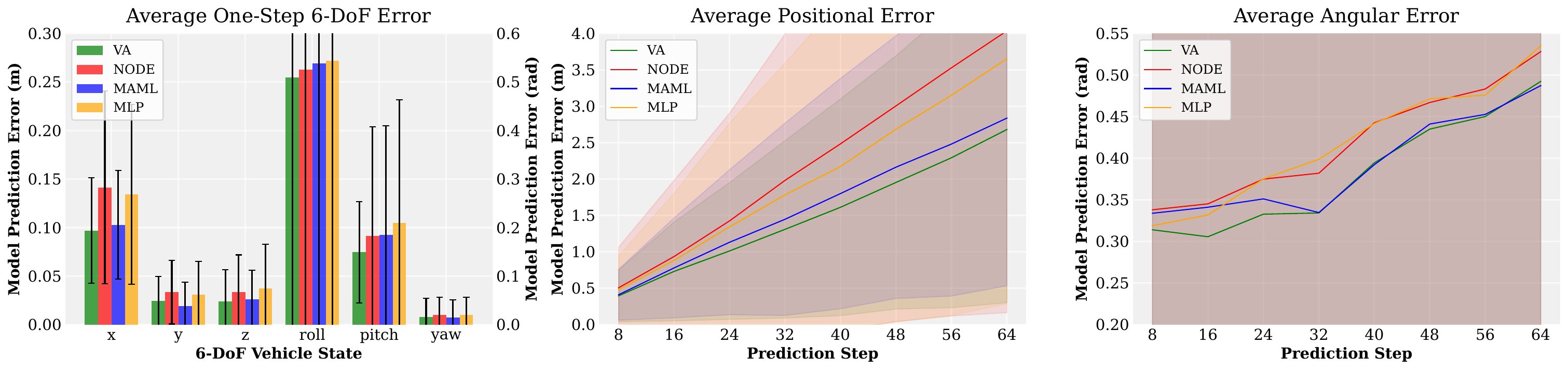}
    \caption{Model Prediction Error of VA and three Baselines: Average One-Step 6-DoF Positional and Angular Error (Left); Prediction Error vs. Prediction Step (Middle and Right).}
    \label{fig::VA_acc}
    \vspace{-10pt}
\end{figure*}

Since the mobile robot’s kinodynamics are invariant under translation and rotation, we adopt a gravity-aligned body frame to enhance data efficiency and improve model accuracy. For each training sample, current state is denoted as:
$[0, 0, 0, \phi_t, \varphi_t, 0, \mathbf{e}_{\text{elev}}, \mathbf{e}_{\text{sem}}] \in \mathbb{R}^{22}
$, 
where the position components $(x, y, z)$ and yaw $\psi$ are zeroed (and therefore omited), while roll $\phi_t$ and pitch $\varphi_t$ retain their world-frame values. The terrain embeddings $\mathbf{e}_{\text{elev}} \in \mathbb{R}^{8}$ and $\mathbf{e}_{\text{sem}} \in \mathbb{R}^{8}$ are derived from the SWAE encoder and further compressed by a 3-layer MLP. The corresponding next state represents the change in vehicle pose:
$
[\Delta x, \Delta y, \Delta z, \phi_{t+1}, \varphi_{t+1}, \Delta\psi, \mathbf{e}_{\text{elev}}, \mathbf{e}_{\text{sem}}] \in \mathbb{R}^{22}
$, 
where $(\Delta x, \Delta y, \Delta z, \Delta \psi)$ represent the body-frame relative motion, and $(\phi_{t+1}, \varphi_{t+1})$ are the absolute roll and pitch angles at the next timestep. We omit $\mathbf{e}_{\text{elev}}$ ad $\mathbf{e}_{\text{sem}}$ in the prediction since they can be acquired from perception.

The vehicle executes sinusoidal random exploration across all environments to collect training data. Steering follows $\mathbf{u}_{\text{steer}}(t)=\sin(\omega_s t)$ with $\omega_s \sim \mathcal{U}(0.1, 0.5)$ Hz, combined with speed $\mathbf{u}_{\text{speed}}(t) = v_c + A\sin(\omega_v t)$, where $\omega_v \sim \mathcal{U}(0.1, 2.5)$ Hz. The speed amplitude $A$ and center velocity $v_c$ are determined such that the minimum speed lies in $\mathcal{U}(0, 2)$ m/s and maximum speed is 3 m/s. This exploration strategy ensures diverse vehicle-terrain interactions, providing rich datasets for training the VA framework.

\subsection{Model Architecture}
Each neural ODE basis function $g_i$ is a 4-layer MLP with $\{24, 313, 313, 313, 313, 6\}$ neurons and ReLU activations. The network takes a concatenated state–action vector as input and outputs a 6-DoF state derivative. The hidden dimension is scaled with the number of basis functions as $h = \lfloor 64\sqrt{k} \rfloor$ to balance model capacity and computational cost. For coefficient computation, we employ a memory-efficient batched least-squares solver that processes example sets in mini-batches of size 256. This design enables handling large datasets while maintaining numerical stability. A regularization term of $\lambda = 10^{-3}$ is applied to improve the numerical stability of the Gram matrix inverse in Eqn.~\eqref{eq:coef}.

The model is trained using multi-step rollout with gradient accumulation to balance memory efficiency and training stability. As shown in Table~\ref{tab:params}, the cosine annealing schedule gradually decays the learning rate from $10^{-3}$ to $10^{-5}$ over 1000 training steps, promoting stable convergence and fine-grained parameter updates.

\subsection{Baselines}
VA is compared against three baseline methods: MLP, first-order MAML, and Neural ODE (NODE).

MLP employs a standard feedforward neural network with the same 4-layer architecture as VA's basis functions. For online adaptation, we fine-tune only the last layer to reduce computational overhead while retaining adaptation capability. Given example data from a new terrain, the model performs 40,000 gradient steps until convergence using the Adam optimizer with a learning rate $5\times10^{-3}$, minimizing the Mean Squared Error (MSE) between predicted and ground-truth state changes until convergence. 

First-order MAML is designed to learn initialization parameters that enable rapid adaptation. During offline training, the model learns across diverse terrain distributions using the standard MAML objective, allowing it to quickly adapt to new conditions. For adaptation, the meta-learned parameters are fine-tuned for 20,000 gradient steps to converge with an inner learning rate $5\times10^{-3}$. However, gradient-based updates remain computationally expensive and too slow to support real-time model inference in navigation systems such as MPPI.

To overcome this challenge, Neural ODE shares the same network architecture but integrates its outputs using the RK4 numerical method. By modeling kinodynamics in continuous time, Neural ODE provides smoother state transitions and more stable predictions. During offline training, the model learns to predict state derivatives that are integrated over a fixed time step of $\Delta t = 0.1$s. For online adaptation, all network parameters are fine-tuned with only 500 gradient steps to converge. The reduced adaptation steps, compared to MLP and MAML, reflect the improved sample efficiency from continuous-time kinodynamic modeling, However, the method still relies on iterative optimization, limiting real-time performance, while VA only uses a least-squares solver. 

%% file: Contents/results.tex
\section{Experiments}

We compare kinodynamics accuracy and simulated navigation performance in Verti-Bench~\cite{xu2025verti}, contrasting  VA against the three baselines. Additionally, we compare kinodynamic predictions of VA and MAML using real-world data collected from a physical 1/10th-scale open-source Verti-4-Wheeler robot~\cite{datar2024toward} in an off-road testbed. 
% in the controllable indoor testbed Verti-Arena\cite{chen2025verti}.

\begin{figure*}[ht]
    \centering
    \includegraphics[width=2\columnwidth]{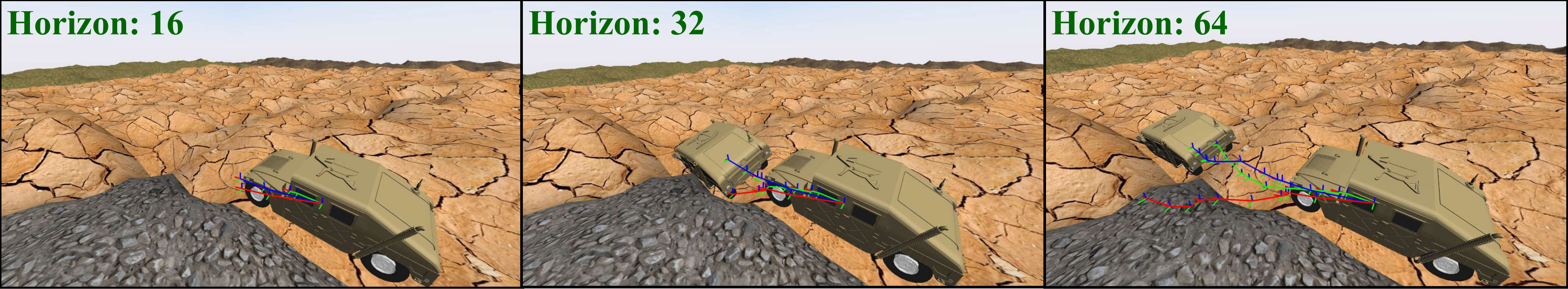}
    \caption{6-DoF Vehicle Trajectories of \textcolor{darkgreen}{VA}, \textcolor{red}{MAML}, and \textcolor{blue}{Ground Truth} with Increasing Horizon: \textcolor{darkgreen}{VA} matches \textcolor{blue}{Ground Truth} even with a long horizon, while \textcolor{red}{MAML} diverges more.}
    \label{fig::VA_traj}
    \vspace{-10pt}
\end{figure*}

\subsection{Prediction Accuracy}
We evaluate the accuracy of VA against all baselines in unseen off-road environments with high elevation level. As shown in Fig.~\ref{fig::VA_acc} left, VA achieves the lowest one-step prediction errors for most 6-DoF state components, with an overall mean error of 0.1367 compared to Neural ODE (0.1561), MAML (0.1476), and MLP (0.1626). In particular, VA outperforms all baselines in rotational states, reducing roll and pitch error by 3.2\% and 22.3\% compared to Neural ODE. For translational states, VA achieves competitive accuracy, with the lowest error in the $x$ direction (0.0969) and $z$ direction (0.0243) relative to the baselines. These results indicate that VA is effective at modeling complex rotational dynamics while remaining robust for translational motions.

Moreover, VA demonstrates lower prediction variance, reflecting more consistent and reliable performance across diverse terrain conditions. As shown in Fig.~\ref{fig::VA_acc} middle and right, when predictions are rolled out over time, all baselines experience rapid error growth and increasing uncertainty, especially for positional states. In contrast, VA maintains stable and accurate long-term predictions, with slower error accumulation and smaller uncertainty bounds. 

Building on these findings, we further evaluate VA’s performance across different terrain difficulty levels. The results in Table~\ref{tab:eva_res} compare the proposed VA framework with three baselines, Neural ODE, MAML, and MLP, across low, medium, and high elevation levels.

% \begin{table}[h!]
% \centering
% \caption{Baseline evaluations across three terrain difficulty levels.}
% \label{tab:eva_res}
% \renewcommand{\arraystretch}{1.4} % Better vertical spacing
% \setlength{\tabcolsep}{3pt}       % Adjust horizontal spacing
% \small
% \begin{tabular}{lcccccc}
% \toprule[1pt]
% \multirow{2}{*}{\textbf{Method}} & 
% \multicolumn{3}{c}{\textbf{MSE} $\downarrow$} & 
% \multicolumn{3}{c}{\textbf{Adaptation Time} $\downarrow$} \\ 
% \cmidrule(lr){2-4} \cmidrule(lr){5-7}
% & \textbf{Low} & \textbf{Medium} & \textbf{High} 
% & \textbf{Low} & \textbf{Medium} & \textbf{High} \\ 
% \midrule

% \rowcolor[gray]{.9}
% VA (Proposed) & \textbf{0.164} & \textbf{0.564} & \textbf{1.451} & \textbf{0.239} & \textbf{0.235} & \textbf{0.241} \\

% Neural ODE    & 0.183 & 0.757 & 1.630 
%               & 1.079 & 1.085 & 1.077 \\

% MAML          & 0.445 & 1.446 & 1.484 
%               & 6.999 & 6.944 & 6.865 \\

% MLP           & 0.351 & 0.896 & 2.081 
%               & 15.563 & 15.571 & 15.562 \\

% \bottomrule[1pt]
% \end{tabular}
% \end{table}

\begin{table}[h!]
\centering
\caption{Baseline evaluations across three terrain difficulty levels.}
\label{tab:eva_res}
\renewcommand{\arraystretch}{1.4} % Better vertical spacing
\setlength{\tabcolsep}{2.5pt}       % Adjust horizontal spacing
\small
\begin{tabular}{lcccccc}
\toprule[1pt]
\multirow{2}{*}{\textbf{Method}} & 
\multicolumn{3}{c}{\textbf{MSE} $\downarrow$} & 
\multicolumn{3}{c}{\textbf{Adaptation Time} $\downarrow$} \\ 
\cmidrule(lr){2-4} \cmidrule(lr){5-7}
& \textbf{Low} & \textbf{Medium} & \textbf{High} 
& \textbf{Low} & \textbf{Medium} & \textbf{High} \\ 
\midrule

\rowcolor[gray]{.9}
VA (Proposed) & \textbf{0.177} & \textbf{0.760} & 2.161 & \textbf{0.309} & \textbf{0.306} & \textbf{0.311} \\

Neural ODE    & 0.200 & 0.999 & 3.128 
              & 1.715 & 1.737 & 1.734 \\

MAML          & 0.186 & 0.814 & \textbf{2.093} 
              & 11.175 & 11.366 & 10.616 \\

MLP           & 0.214 & 0.904 & 2.214 
              & 40.889 & 40.556 & 40.777 \\

\bottomrule[1pt]
\end{tabular}
\end{table}

The VA demonstrates superior performance in both accuracy and efficiency. In terms of prediction accuracy, VA achieves the lowest MSE on low (0.177) and medium (0.760) elevation levels, outperforming Neural ODE by 11.5\% and 23.9\%, MAML by 4.8\% and 6.6\%, and MLP by 17.3\% and 15.9\%, respectively. On high elevation level, VA achieves competitive performance (2.161) with only a 3.2\% higher MSE compared to the best-performing MAML (2.093), while still outperforming Neural ODE by 30.9\% and MLP by 2.4\%. These results highlight VA's consistent ability to model complex vehicle-terrain interactions across varying terrain difficulty levels.

In terms of adaptation time, VA adapts the fastest, requiring approximately 0.31s, which is over 5X faster than Neural ODE ($\sim$1.73s) and more than 30X faster than MAML ($\sim$11s) and MLP ($\sim$40s). This highlights VA's ability to achieve both high accuracy and rapid adaptation, effectively balancing precision and efficiency.

Fig.~\ref{fig::VA_traj} compares the predicted trajectories from VA and MAML against the ground truth across different prediction horizons in the most challenging high elevation terrain. At horizon 16, both methods closely follow the ground truth. By horizon 32, while both VA and MAML generally maintain the correct direction, MAML begins to drift to the left and VA slightly to the right. By horizon 64, MAML accumulates substantial error, with its trajectory extending onto rocky terrain on the left. In contrast, VA continues to closely track the ground truth.

\subsection{Ablation Studies}
We conduct ablation studies to evaluate the contribution of each key component in VA: (i) elevation embeddings and (ii) semantic embeddings. Each variant removes one of these components from the full VA framework:
\begin{itemize}
    \item \textbf{VA}: The complete VA framework;

    \item \textbf{VA w/o semantic}: Removes the semantic embeddings. The model uses only the 6-DoF pose and elevation information, capturing geometric terrain features but lacking knowledge about terrain semantics such as friction or deformability;
    
    \item \textbf{VA w/o elevation}: Removes the elevation embeddings. In this variant, the model relies solely on semantic embeddings and the 6-DoF pose to capture terrain interaction, without direct geometric information about terrain elevation changes; and
    
    \item \textbf{VA w/o both}: Uses only the 6-DoF pose without elevation or semantic embeddings. This baseline captures purely kinodynamic behavior without terrain context, relying solely on vehicle motion states.
\end{itemize}

\begin{table}[h!]
\centering
\caption{Ablation Study: VA MSE across three terrain difficulty levels.}
\label{tab:va_abl}
\renewcommand{\arraystretch}{1.4} % Slightly tighter vertical spacing
\setlength{\tabcolsep}{10pt} % Reduce horizontal spacing
\small
\begin{tabular}{lccc}
\toprule[1pt]
\multirow{2}{*}{\textbf{Variant}} & 
\multicolumn{3}{c}{\textbf{MSE} $\downarrow$} \\ 
\cmidrule(lr){2-4} 
& \textbf{Low} & \textbf{Medium} & \textbf{High} \\ 
\midrule
\rowcolor[gray]{.9}
\textbf{VA} (Proposed) & 
\textbf{0.177} & 
\textbf{0.760} & 
\textbf{2.161} \\

\textbf{VA} w/o semantic & 
0.201 & 
0.761 & 
2.292 \\

\textbf{VA} w/o elevation & 
0.220 & 
0.840 & 
2.559 \\

\textbf{VA} w/o both & 
0.177 & 
0.761 & 
3.038 \\

\bottomrule[1pt]
\end{tabular}
\end{table}

The results in Table~\ref{tab:va_abl} underscore the importance of both elevation and semantic embeddings within the VA framework across terrain difficulty levels. On low elevation terrain, the complete VA framework achieves the lowest MSE (0.177), outperforming variants w/o semantic (0.201) and elevation (0.220) embeddings. Interestingly, the VA w/o both also achieves an MSE of 0.177, indicating that for low elevation level, basic kinodynamic modeling using only the 6-DoF pose is sufficient to achieve comparable accuracy. As terrain complexity increases, the impact of elevation embeddings becomes more pronounced. On medium elevation terrain, removing elevation information increases MSE by 10.5\%, while removing semantic embeddings causes only a negligible 0.1\% increase, showing that elevation data is key for capturing geometric variations. On high elevation terrain, all ablations show clear performance degradation: removing semantic embeddings increases error by 6.1\%, while removing elevation embeddings causes a much larger 18.4\% increase. The VA w/o both suffers the most, with a 40.5\% increase in error compared to the full VA framework. These findings demonstrate that elevation embeddings are critical for geometric terrain understanding, semantic embeddings provide complementary insights for surface properties, and their combination is essential for robust vehicle-terrain interaction modeling in challenging off-road environments.

\subsection{Simulated Navigation Performance}
VA and the best baseline MAML are integrated into the MPPI planner, whose navigation performance is evaluated across terrain difficulty levels using three metrics: (i) number of successful trials (out of 5), (ii) mean traversal time (of successful trials in seconds), and (iii) average roll/pitch angles with variance (in degrees). We also compare VA adapted once per new environment (VA$^{*}$) and adapted every 5s (VA$^{\dagger}$), as well as MAML adapted once (MAML$^{*}$), since the significant computation disallows MAML do adapt online. 

\begin{table}[h!]
\caption{\textbf{Simulated Navigation Performance:} Success Rate, Traversal Time, Roll, and Pitch. $^{*}$: adapted once, $^{\dagger}$: adapted every 5s.}
\label{tab:sim}
\centering
\renewcommand{\arraystretch}{1.0} % Better vertical spacing
\setlength{\tabcolsep}{5pt}       % Adjust horizontal spacing
\small
\begin{tabular}{cccc} % Only 3 columns needed
\toprule[1pt]
Low Elevation & \textbf{VA$^{*}$} & \textbf{VA$^{\dagger}$} & \textbf{MAML$^{*}$} \\
\midrule
Success Rate $\uparrow$      & \textbf{5/5}                    & \textbf{5/5} & 4/5 \\
Traversal Time $\downarrow$  & 31.16s & 34.73s & \textbf{28.90s} \\
Roll $\downarrow$            & 
2.36\textdegree$\pm$0.06\textdegree &
\textbf{2.23\textdegree$\pm$0.05\textdegree}   &  2.61\textdegree$\pm$0.11\textdegree \\
Pitch $\downarrow$           & 
2.17\textdegree$\pm$0.04\textdegree &
1.98\textdegree$\pm$0.04\textdegree   &  \textbf{1.97\textdegree$\pm$0.04\textdegree} \\
\midrule
Medium Elevation & \textbf{VA$^{*}$} & \textbf{VA$^{\dagger}$} & \textbf{MAML$^{*}$}  \\
\midrule
Success Rate $\uparrow$    & \textbf{3/5}  & 2/5 & 0/5 \\
Traversal Time $\downarrow$ & \textbf{33.94s} & 39.98s & - \\
Roll $\downarrow$            & 
5.73\textdegree$\pm$0.25\textdegree &
\textbf{5.06\textdegree$\pm$0.62\textdegree}   & 
- \\
Pitch $\downarrow$           & 
4.74\textdegree$\pm$0.07\textdegree &
\textbf{4.17\textdegree$\pm$0\textdegree}   & 
- \\
\midrule
High Elevation & \textbf{VA$^{*}$} & \textbf{VA$^{\dagger}$} & \textbf{MAML$^{*}$}  \\
\midrule
Success Rate $\uparrow$     & \textbf{4/5} & 1/5 & 0/5 \\
Traversal Time $\downarrow$ & 32.75s & \textbf{32.29s} & - \\
Roll $\downarrow$          & 9.18\textdegree$\pm$1.34\textdegree  & \textbf{5.83\textdegree$\pm$-} & - \\
Pitch $\downarrow$     & \textbf{5.91\textdegree$\pm$0.17\textdegree}      & 6.24\textdegree$\pm$- & - \\
\bottomrule[1pt]
\end{tabular}
\end{table}
\vspace{-3pt}

Table~\ref{tab:sim} shows that our single adaptation method, VA$^{*}$, achieves the best overall performance, consistently surpassing MAML$^{*}$ in success rate. Surprisingly, VA$^{\dagger}$ performs worse for medium and high elevation levels than VA$^{*}$. This is because VA$^{*}$ computes a single set of coefficients that represent an averaged kinodynamic model after seeing data sampled from the entire new environment. However, VA$^{\dagger}$ continuously updates its coefficients based on recent trajectory history, which can lead to mismatch when terrain types change significantly. As a result, the updated coefficients may work well for the previous terrain but become poorly suited when the vehicle encounters entirely new semantic terrain types. These findings indicate that, while our one-time adaptation strategy provides greater robustness and reliability, future work should focus on refining recursive updates to better handle rapid and frequent terrain changes.

\subsection{Physical Experiment for Sim-to-Real Adaptation}
Trained with simulation data from Verti-Bench~\cite{xu2025verti}, we adapt VA and MAML using physical data collected on real-world, vertically challenging, off-road terrain to test their sim-to-real adaptation. 
The real-world dataset includes driving a Verti-4-Wheeler robot through diverse rocky terrain in the off-road testbed, as shown in Fig.~\ref{fig::motivation}, for 70.3 seconds. We use 28.1 seconds of data for adaptation and evaluate and compare the kinodynamic modeling accuracy of VA and MAML with the rest 42.2 seconds of data. 

\begin{table}[ht]
\centering
\caption{Physical Experiment Results}
\label{tab:testbed_res}
\renewcommand{\arraystretch}{1.4}
\setlength{\tabcolsep}{3pt} % Adjust column spacing
\small
\begin{tabular}{lcccccc}
\toprule[1pt]
\multirow{2}{*}{\textbf{Method}} &
\multicolumn{6}{c}{\textbf{MSE} $\downarrow$} \\
\cmidrule(lr){2-7}
& \textbf{x} & \textbf{y} & \textbf{z} & \textbf{roll} & \textbf{pitch} & \textbf{yaw} \\
\midrule
\rowcolor[gray]{.9}
VA (Proposed) \rule{0pt}{3.0ex} & \textbf{0.014} & \textbf{0.010} & \textbf{0.002} & \textbf{0.066} & \textbf{0.041} & \textbf{0.024} \\[1ex]
MAML \rule{0pt}{3.0ex} & 0.016 & 0.201 & 0.003 & 0.068 & 0.059 & 0.025 \\
\bottomrule
\end{tabular}
\end{table}
\vspace{-5pt}

The physical experiment results in Table~\ref{tab:testbed_res} demonstrate that VA achieves more accurate predictions than MAML across all DoFs, especially in lateral motion. In particular, VA significantly reduces the y-direction error by 95\% from 0.201 to 0.010, indicating a much stronger ability to model lateral vehicle kinodynamics on vertically challenging terrain. Both methods perform similarly in vertical motion (z) and yaw, but VA achieves lower errors in translational states (x, y) and orientation states (roll, pitch). These results confirm that VA can enable  terrain-aware, sim-to-real adaptation, improves real-world prediction accuracy, and provides a more reliable foundation for navigation in complex off-road environments.

% \begin{figure}[h!]
%     \centering
%     \includegraphics[width=\columnwidth]{example-image-a}
%     \caption{Physical Off-Road Testbed Similar to Verti-Bench.} 
%     \label{VA::real}
% \end{figure}

%% file: Contents/conclusion.tex
\section{Conclusions and Limitations}
We present VA, a novel online adaptation framework for terrain-aware kinodynamic modeling in off-road environments. By integrating elevation and semantic embeddings, VA enables rapid adaptation to unseen terrain through efficient least-squares optimization. Our evaluation demonstrates that VA improves prediction accuracy by up to 23.9\% while achieving 5X faster adaptation than baselines, effectively capturing complex vehicle-terrain interactions.

A key limitation of this work is that our adaptation method updates coefficients once, which may restrict its ability to handle rapidly changing terrain conditions. In future work, we plan to explore more robust recursive approaches, such as recursive least squares, to enable continuous and stable updates. In addition, elevation and semantic embeddings are currently combined through simple concatenation, which may not fully capture rich cross-modal relationships. To address this, we aim to investigate advanced fusion techniques, such as attention mechanisms or cross-modal transformers, to better integrate geometric and semantic information and further enhance adaptation performance.

\section*{Acknowledgments}
This work has taken place in the RobotiXX Laboratory at George Mason University. RobotiXX research is supported by National Science Foundation (NSF, 2350352), Army Research Office (ARO, W911NF2320004, W911NF2520011), Army Ground Vehicle Systems Center (GVSC), Google DeepMind (GDM), Microsoft Research (MSR), Clearpath Robotics, FrodoBots Lab, Raytheon Technologies (RTX), Tangenta, 4-VA, Mason Innovation Exchange (MIX), and Walmart.